**TITLE PAGE**

**Title**

# Automatic sleep stage classification with deep residual networks in a mixed-cohort setting

**Authors and affiliations**

Alexander Neergaard Olesen[1,2,3], Poul Jennum[3*], Emmanuel Mignot[2*], Helge Bjarup Dissing Sørensen[1*],

[1]Department of Health Technology, Technical University of Denmark, Kgs. Lyngby, Denmark

[2]Stanford Center for Sleep Sciences and Medicine, Stanford University, Palo Alto, CA, USA

[3]Danish Center for Sleep Medicine, Department of Clinical Neurophysiology, Rigshospitalet, Glostrup, Denmark

*These authors have contributed equally

**Corresponding author**

Alexander Neergaard Olesen, aneol@dtu.dk

Department of Health Technology, Technical University of Denmark

Ørsteds Plads, building 349, room 010

2800 Kgs. Lyngby, Denmark

**Institution where work was performed**

Stanford Center for Sleep Sciences and Medicine, Stanford University

Department of Health Technology, Technical University of Denmark

**ABSTRACT**

**Study Objectives:** Sleep stage scoring is performed manually by sleep experts and is prone to subjective interpretation of scoring rules with low intra- and interscorer reliability. Many automatic systems rely on few small-scale databases for developing models, and generalizability to new datasets is thus unknown. We investigated a novel deep neural network to assess the generalizability of several large-scale cohorts.

**Methods:** A deep neural network model was developed using 15.684 polysomnography studies from five different cohorts. We applied four different scenarios: 1) impact of varying time-scales in the model; 2) performance of a single cohort on other cohorts of smaller, greater or equal size relative to the performance of other cohorts on a single cohort; 3) varying the fraction of mixed-cohort training data compared to using single-origin data; and 4) comparing models trained on combinations of data from 2, 3, and 4 cohorts.

**Results:** Overall classification accuracy improved with increasing fractions of training data (0.25%: 0.782 ± 0.097, 95% CI [0.777 – 0.787]; 100%: 0.869 ± 0.064, 95% CI [0.864 – 0.872]), and with increasing number of data sources (2: 0.788 ± 0.102, 95% CI [0.787 – 0.790]; 3: 0.808 ± 0.092, 95% CI [0.807 – 0.810]; 4: 0.821 ± 0.085, 95% CI [0.819 – 0.823]). Different cohorts show varying levels of generalization to other cohorts.

**Conclusions:** Automatic sleep stage scoring systems based on deep learning algorithms should consider as much data as possible from as many sources available to ensure proper generalization. Public datasets for benchmarking should be made available for future research.

## STATEMENT OF SIGNIFICANCE

Manual annotation of polysomnography studies is subject to human bias with multiple studies showing variations in how sleep experts score sleep. Most research in automatic sleep stage classification models use small-scale data from a single origin, and it is unknown how these models generalize to new data. We developed an algorithm for automatic scoring of sleep stages using raw polysomnography data and obtain state-of-the-art classification performance on a large number of test subjects. Our algorithm was tested under different conditions to compare generalizability. We found that using data from many different sources improves classification performance, and that models trained on single-origin data generalize inconsistently to new data. Future researchers should take multiple datasets into account when developing sleep scoring models.

## INTRODUCTION

Sleep staging is important to the analysis of human sleep with about 845,000 sleep studies performed in 2014 in the US alone[1]. Briefly, a standard clinical sleep study consists of a full-night polysomnography (PSG) comprising electroencephalography (EEG), electrooculography (EOG), electromyography (EMG), electrocardiography (ECG), thoraco-abdominal inductance plethysmography, oronasal thermal flow, nasal pressure, and blood saturation recordings. These studies are then evaluated by experts for the presence of events of clinical relevance, as determined by standards created by the American Academy of Sleep Medicine (AASM), such as the number of blood oxygen desaturations, micro-arousals, leg movements, periods of cessated breathing, etc. Furthermore, the overall sleep architecture is captured in a hypnogram conducted by labeling every 30 s of PSG data into one of five stages of sleep: wakefulness (W), rapid eye movement (REM) sleep, and non-REM stage 1, 2, and 3 (N1, N2, N3). The latter three stages are distinguished by distinct EEG amplitude and frequency distributions, the presence of specific EEG micro-events and arousability differences reflecting sleep depth. Sleep stage labeling is summarized in key metrics, such as the percentage of total sleep time (TST) spent in any of the five stages (%W, or wake after sleep onset, WASO; %REM; %N1; %N2; %N3), and visually in the form of a hypnogram, which shows temporal progression of sleep stages across the night. Current clinical practice (gold standard) of sleep study analysis is manual scoring and annotation of sleep stages and sleep events based on guidelines from the AASM[2]. These guidelines, based on observations made in healthy young males almost 70 years ago are problematic for several reasons: a) technicians will never score the same data the exact same way as another technician, or even the same way twice[3–7]; b) normal sleep from healthy young males may not reflect sleep patterns of patients referred to sleep clinics; and c) the 30 s epoch rule is arbitrary and was based on physical limitations of recording equipment when PSGs were recorded on paper.

Automatic sleep stage classification has not yet seen wide-spread adoption in clinical practice despite ongoing research demonstrating feasibility and industrial interests[8]. A major issue has been a lack of available data for designing and training models. The publicly available PhysioNet Sleep-EDF and the expanded version[9,10] has been used extensively for training both shallow and deep learning-based machine learning models[11–13], but given its small sample size and homogeneity (most papers use the same healthy 20 subjects), it is questionable how well models derived from this data generalize to unseen data, even if high classification performance is often reported[8]. Other databases which have been extensively used include the St. Vincent's University Hospital and University College Dublin Sleep Apnea Database ($n = 25$)[9,14], and the Montreal Archive of Sleep Studies (MASS, $n = 200$)[13,15–19]. The argument for using deep learning-based models to classify high-dimensional electrophysiological data, e.g. PSGs, into discrete outcomes such as sleep stages is compelling, because of their ability to capture variability in the underlying, highly complex, data representations, that might be missed by  machine learning methods relying on manual feature engineering. In the image, speech, and natural language processing domains, the success of deep learning models using untransformed data have been unsurpassed in the last decade, thanks largely due to the availability of ever-increasing amounts of compute resources and more significantly very large, robust and diverse datasets[20].

Recently deep learning models for automatic sleep stage classification have been developed and validated using two or more databases or cohorts[21–23], or using a single large volume cohort[22,24,25]. The assumption has been that by incorporating multiple sources of variance in the dataset used for training (e.g. from multiple technicians, sites, recording setups, equipment, etc.), final models will be better at generalizing to new, unseen data. However, no study to date has investigated multiple, large-scale cohorts for automatic sleep stage classification, or how different cohorts generalize to one another.

In this work, we describe a deep learning-based sleep stage classification algorithm trained and validated on raw PSG data from multiple, large-scale cohorts for a total of 15,684 studies, that outputs a probability distribution over all sleep stages at a given time resolution. Considering the amount of data available, our aim was to evaluate: 1) how well does performance of individual cohorts generalize to others; 2) how much data is needed for accurate sleep staging; 3) how many cohorts are necessary for that same goal; and 4) which is better, more data, or more diverse data. To our knowledge, this is one of the largest, if not the largest, study on automatic sleep stage classification in terms of PSG volume and diversity.

## METHODS

### Cohort descriptions

To investigate and conclude on generalizability of any machine learning or sleep stage classification model, multiple heterogenous datasets must be used for training, validation and testing purposes. In this work, we collected datasets from five different sources, each dataset containing a diverse collection of subjects presenting with multiple disease phenotypes. Details of the separate cohorts are shown in Table 1 along with reported *p*-values highlighting cohort differences. Each cohort was split into a training, validation and testing *subset* in proportions of 87.5%, 2.5% and 10%, respectively, using random sampling without replacement among unique subjects, so that no subject is shared between subsets. With these percentages, we maximize the number of PSGs available for training, while still reserving enough PSGs for validation and testing. Collecting all the separate subsets across cohorts forms a training, validation, and testing *partition*, containing the respective subsets from all five cohorts.

### Institute of Systems and Robotics, University of Coimbra Sleep Cohort (ISRUC)

This cohort contains 126 recordings from 118 unique subjects recorded at the Sleep Medicine Centre of the Hospital of Coimbra University, Portugal, in the period 2009–2013[26]. The cohort comprises three subgroups: subgroup I contains 100 PSGs of subjects with diagnosed sleep disorders, generally sleep apnea; subgroup II contains 16 recordings of eight subjects most of which are also diagnosed with sleep apnea; and subgroup III contains recordings from 10 subjects with no diagnosed sleep disorders. All PSGs were recorded with the same recording hardware and software and each was scored by two technicians for sleep stages and sleep events according to the AASM guidelines. ISRUC-Sleep is a freely accessible resource and all data and PSG files can be located at https://sleeptight.isr.uc.pt/ISRUC_Sleep/.

### The MrOS Sleep Study (MrOS)

The MrOS sleep study is part of the larger Osteoporotic Fractures in Men Study, which aims to understand the relationships between sleep disorders, fractures, and vascular diseases in community-dwelling men[27–29]. It consists of 2,907 in-home PSG recordings with an additional 1,026 follow-up PSG studies from subjects recruited from six different clinical centers in the USA. Each recording was annotated by an expert technician according to Rechtschaffen and Kales (R&K) criteria for sleep staging[30]. For compatibility with AASM guidelines, we combined stages labeled S3 and S4 into N3. All data were accessed from the National Sleep Research Resource (NSRR) repository[31,32].

### The Sleep Heart Health Study (SHHS)

The SHHS is a large, multi-center study on cardiovascular outcomes related to sleep disorders with a specific focus on sleep-disordered breathing[33,34]. The cohort consists of 6,441 subjects above 40 years old recruited between 1995 and 1998 undergoing in-home PSG (SHHS Visit 1) with subsequent follow-up PSG between 2001 and 2003 in 3,295 subjects (SHHS Visit 2). PSG recordings were annotated for sleep stages by trained and certified technicians according to R&K rules. From the original cohort we extracted 5,793 PSGs and annotations from Visit 1, and 2,651 from Visit 2, and aggregated S3 and S4 stages into N3 similar to MrOS. All data were accessed from NSRR repository.

**Wisconsin Sleep Cohort (WSC)**

WSC is a population-based study of sleep-disordered breathing in government workers in Wisconsin, USA that was initiated in 1988[35,36]. In this work, we used 2412 PSGs from 1091 unique subjects in the WSC sample scored by expert technicians according to R&K rules with subsequent merging of S3 and S4 into N3.

**Stanford Sleep Cohort (SSC)**

PSGs from this cohort originate from patients referred for sleep disorders evaluation and recorded at the Stanford Sleep Clinic since 1999. The specific sample used in this study represents a small subset ($n$ = 772) of the whole cohort, which was selected and described in detail in previous studies[37,38] scored according to R&K or AASM guidelines according to prevailing standard at the time of evaluation.

**Signal pre-processing pipeline**

Electrophysiological signals corresponding to the minimum acceptable montage for sleep staging available across all cohorts were extracted for each PSG. These included a central EEG (either C3 or C4 referenced to the contra-lateral mastoid), left and right EOG referenced to the contra-lateral mastoid, and a single submentalis EMG. The choice between C3 and C4 was determined based on the lowest total signal energy across the entire duration of the PSG to avoid excessive signal popping. Other methods to determine appropriate channels include algorithms based on shortest Mahalanobis distance to an already determined reference distribution[21], but was not investigated in this study. All signals were resampled to $f_s$ = 128 Hz using a polyphase filtering procedure irrespective of original sampling frequency; and subsequently filtered using a zero-phase approach with 4$^{th}$ order Butterworth IIR filters (0.5 to 35 Hz band pass for EEG and EOG; 10 Hz high pass for EMG) in accordance with AASM filter specifications[2]. Each signal was normalized to zero mean and unit variance to accommodate differences in recording equipment and baselines; and to compress the dynamic range into something easily trainable for the neural network architecture. We denote by $C$ the number of input signals supplied to the neural network, where in this case $C = 4$.

**Machine learning problem**

We designate by $\mathcal{X} \in \mathbb{R}^{C \times T}$ the set of 30 s input data segments with $C$ input channels and segment length $T$, and the corresponding classifications by $\mathcal{Y} = \{y \in \mathbb{R}_+^K \mid \sum_i y_i = 1\}$, where $K = 5$ corresponds to the five sleep stages. Thus, $y$ is a probability simplex, which maps to the ordered set $\mathcal{S} = \{\mathrm{W}, \mathrm{N1}, \mathrm{N2}, \mathrm{N3}, \mathrm{REM}\}$ by the argmax function such that argmax $y : \mathcal{Y} \rightarrow \mathcal{S}$. Furthermore, as we are potentially interested in classifying multiple sleep stages at once, we extend the problem of classifying a single sleep stage given $x \in \mathcal{X}$ to a sequence-to-sequence problem, in which we desire to learn a differentiable function representation $\Phi$, that maps a sequence of 30 s epochs $\mathbf{x} \in \mathbb{R}^{C \times \alpha T}$ to their corresponding label probabilities $\mathbf{y} \in \mathbb{R}^{K \times \alpha}$, where $\alpha$ is a parameter that controls the sequence length. If e.g. $\alpha = 8$, the sequence x contains 4 min of successive PSG data described by 8 epochs of length 30 seconds. Furthermore, we denote by $[\![a, b]\!]$ the set of integers from $a$ to $b$, i.e. $[\![a, b]\!] \equiv \{n \in \mathbb{N} | a \leq n \leq b\}$, and by $[\![N]\!]$ the shorthand form of $[\![1, N]\!]$.

**Network architecture**

As the representation of $\Phi$, we adapted and extended a previously published neural network architecture for automatic sleep stage classification, which was based on a variant of the ResNet-50 architecture commonly used for two-dimensional image classification tasks, but adapted and re-trained from scratch for the specific use-case of one-dimensional, time-dependent signals in the PSG[24]. This network has the advantage that it does not require any manual feature engineering and extraction compared to previous state of the art sleep stage classification models[21]. An overview of the proposed network architecture is provided graphically in Figure 1 and Table 2. Briefly, the architecture consists of four modules:

1) an initial mixing module $\varphi_{\mathrm{mix}} : \mathbb{R}^{1 \times C \times T} \rightarrow \mathbb{R}^{C \times 1 \times T}$
2) a feature extraction module $\varphi_{\mathrm{feat}} : \mathbb{R}^{C \times 1 \times T} \rightarrow \mathbb{R}^{f_0 2^{R+1} \times 1 \times T/2^R}$
3) a temporal processing module $\varphi_{\mathrm{temp}} : \mathbb{R}^{f_0 2^{R+1} \times 1 \times T/2^R} \rightarrow \mathbb{R}^{2n_h \times T/2^R}$, and
4) a classification module $\varphi_{\mathrm{clf}} : \mathbb{R}^{2n_h \times T/2^R} \rightarrow \mathbb{R}^{K \times T/2^R}$.

Thus, we obtain a differentiable representation of the function $\Phi : \mathbb{R}^{C \times T} \rightarrow \mathbb{R}^{K \times T/2^R}$ as

$$\Phi(\mathbf{x}) = \varphi_{\mathrm{clf}}\left(\varphi_{\mathrm{temp}}\left(\varphi_{\mathrm{feat}}\left(\varphi_{\mathrm{mix}}(\mathbf{x})\right)\right)\right).$$

The output of this function is the matrix $\mathbf{y} \in \mathbb{R}^{K \times T/2^R}$ containing sleep stage probabilities in the sequence of PSG data evaluated every second.

*Mixing module*

The raw input data is input to this module, which encourages non-linear channel mixing similar to what has been proposed in recent literature[16,39–41]. The module is realized using a single 2D convolutional operation outputting $C$ feature maps computed using single-strided $C \times 1$ kernels followed by rectified linear unit (ReLU) activations.

*Feature extraction (residual network) module*

This is comprised of a succession of *R* residual blocks (see Figure 1), which are responsible for the bulk feature extraction from the channel-mixed data. Each residual block is realized using bottlenecks of first a $1 \times 1$ convolution to reduce the number of feature maps, then a $1 \times 3$ convolution and lastly a $1 \times 1$ convolution to finally increase the number of feature maps. Each convolution operation was followed by a batch normalization[42] and ReLU activation except after the last convolutional layer, where shortcut projections are added before the activation[43]. This type of block structure enables the design and training of very deep networks without the risk of vanishing gradients due to the projection shortcuts[44].

*Temporal processing module*

This module is realized by a bidirectional gated recurrent unit (GRU)[45] in order to accommodate temporal dependencies in the PSG. The GRU runs through the temporal dimension of the output from $\varphi_{\text{feat}}$ of $T/2^R$ time steps each containing $f_0 2^{R+1}$ features and outputs $n_h$ features in each direction for each time step. By running both forward and backward, we can accommodate that technicians base their scoring on looking backwards as well as ahead in time in each time segment (typically 30 s).

*Classification module*

The final module in the architecture performs actual classification based on the forward and backward features for each time step outputted from $\varphi_{\text{temp}}$. It is realized by a single convolutional operation with a subsequent softmax activation to compute a probability distribution over the *K* sleep stage classes, such that the probability of sleep stage $i$ at time step $n$ is given by $y_i^{(n)} = \frac{\exp(a_i)}{\sum_k \exp(a_k)}$, where $a_i \in \boldsymbol{a}$ is the activation of the last layer in the network and $k = [\![K]\!]$.

**Loss function**

The network was trained end-to-end with respect to a loss function, that takes the output probabilities from the network $\mathbf{y} = \Phi(\mathbf{x})$ and calculates the loss as

$$\mathcal{L}(\mathbf{y}) = -\sum_{n=1}^{30/\tau} \sum_{k=1}^{K} t_k{}^{(n)} \log\left(y'^{(n)}_k\right), \tag{1}$$

$$y'^{(n)}_k = \frac{1}{\tau} \sum_{i=\tau(j-1)+1}^{\tau n} y^{(i)}_k, \tag{2}$$

which is the cross-entropy between successive time-averaged classifications (parameterized by the number of successive one-second predictions $\tau$), and the ground truth labels $t$ broadcasted to $30/\tau$ labels per 30 s segment. This way, we can acquire predictions every second, that can be combined in time at intervals given by $\tau$.

**Experimental setups**

We set up three different experiments in this study.

A) We wished to investigate the effect of increasing the complexity of the recurrent module by varying the number of units $n_h$ in the module $\varphi_{\text{temp}}$ in the space $n_h = 2^k,\ k \in [\![6,11]\!]$. We hypothesize that there exists a sweet-spot in the number of hidden units that balances computational complexity with classification performance, i.e. classifying a sequence of sleep stage labels given a corresponding sequence of outputs from $\varphi_{\text{feat}}$. The results of this experiment were furthermore used to determine parameters for models in subsequent experiments.

B) Since we have several cohorts at our disposition of both clinical and research origin, we can investigate the compatibility and inherent generalizability of the different cohorts in two ways: 1) we set aside a single cohort for testing, while we train the models on the remaining four (leave-one-cohort-out, LOCO training); and 2) we train on a single cohort, while we set aside the remaining four for testing (leave-one-cohort-in, LOCI training).

C) Generalizability can also be investigated in another way, which can answer the question of how many data sources is necessary. We trained models with all possible 2-, 3-, and 4-combinations of cohorts, i.e. one run trained on ISRUC and MrOS training data, another run with ISRUC and SHHS train data, a third with ISRUC and SSC, etc., with all runs subjected to subsequent evaluation on the test partition.

D) Previous studies have already investigated the performance of automatic sleep staging algorithms using shallow machine learning models. At the time of writing however, none have investigated the effect of available training data for deep learning models at this magnitude (up to tens of thousands). We therefore trained models on 0.25%, 0.5%, 1%, 5%, 10%, 25%, 50%, 75% and 100% of the data available for training. Specifically, some of these fractions of the total number of PSGs correspond roughly to the number of PSGs in the training partitions in each cohort, allowing for direct comparisons between training a model with mixed- and single-cohort training data.

Common for all experiments were the default parameter values $C = 4$, $f_s = 128$ Hz, $T = \tau f_s$, $K = 5$, $R = 7$, and $f_0 = 4$ for the number of input channels, sampling frequency, the sequence length, the number of sleep stages, the number of consecutive residual blocks, and the base filter kernel size, respectively. All models were trained for 50 epochs (passes through the training partition) and the model with the highest Cohen's kappa value on the validation partition was subsequently selected for testing. All models were trained end-to-end with backpropagation using the Adam optimizer[46] with a learning rate of $10^{-4}$, $\beta_1 = 0.9$, and $\beta_2 = 0.999$ to minimize the loss function specified by Eq. (1) and Eq. (2). All network weights and bias terms were initialized using the uniform Glorot initialization scheme[47].

**Performance metrics and model evaluation**

For each experiment we evaluated model performance using the overall accuracy (Acc) and Cohen's kappa (κ) in order to into account the possibility of chance agreement between the model gold standard. Given a confusion matrix $\mathbf{C}$ with element $c_{ij}$ being the number of epochs belonging to sleep stage $i$ but classified to be in sleep stage $j$, we define the overall accuracy for a given model as

$$\mathrm{Acc} = \frac{\sum_{i=j} c_{ij}}{\sum_{i,j} c_{ij}}$$

i.e. the sum of the trace of $\mathbf{C}$ divided by the total count. The Cohen's kappa metric is defined as

$$\kappa = \frac{p_o - p_e}{1 - p_e}$$

where $p_o = \mathrm{Acc}$ is the observed agreement (i.e. accuracy) and $p_e$ is the expected chance agreement, which can be reformulated in terms of the outer product between the row and column sums (class-specific recall and precision) of $\mathbf{C}$.

**Data and source code availability**

All model training and testing code was implemented in PyTorch v. 1.2[48]. Model performances were assessed using custom Python scripts using scikit-learn[49]. Source code and pre-trained models will be made available at https://github.com/neergaard/deep-sleep-pytorch.git and https://github.com/Stanford-STAGES/deep-sleep-pytorch.git upon publication of this paper. Data from ISRUC are publicly available at https://sleeptight.isr.uc.pt/ISRUC_Sleep/, while access to data from MrOS and SHHS can be requested from the NSRR. Anonymized PSG data from SSC including selected demographic data are available at https://stanfordmedicine.app.box.com/s/r9e92ygq0erf7hn5re6j51aaggf50jly.

## RESULTS

In this section we report on the results of the three experiments described in the **Experimental setups** section.

### Temporal context impact on model performance

In Figure 2 we show how the model performance depends on the temporal context and complexity of the temporal processing module, when evaluating the model on the validation partition. Results are further detailed in Table S1. Specifically, we observe a drastic change in Cohen's kappa just by introducing a simple recurrent unit into the network as shown in Figure 2a, where Cohen's kappa increases from $0.645 \pm 0.126$ (95% CI: $[0.633 - 0.657]$) at $n_h = 0$ to $0.720 \pm 0.120$ (95% CI: $[0.709 - 0.731]$) at $n_h = 64$. We did not observe any major changes when increasing the number of hidden units beyond 64, although we did see a maximum Cohen's kappa of $0.734 \pm 0.111$ (95% CI: $[0.723 - 0.744]$) at $n_h = 1024$, which is shown in the inset in Figure 2a. We observed a general increase in Cohen's kappa when classifying longer sequences than 2 min ($0.726 \pm 0.114$, 95% CI: $[0.715 - 0.737]$), but did not see any major differences when classifying over more than 3 min sequences ($0.733 \pm 0.123$, 95% CI: $[0.721 - 0.744]$). Subsequent models were fixed with $n_h = 1024$ corresponding to a sequence length of 5 min.

### Model classifications converge to 30 s predictions given sufficient training data

Furthermore, we analyzed the classification performance of the model given a specific sequence length by looking at the average prediction accuracy across all 5 min sequences in all subject PSGs in the test partition, similar to what Brink-Kjaer et al. has shown previously[50]. In Figure 2c, we show how the average classification accuracy in a 5 min sequence both depends on the amount of data and the frequency of evaluating the model output, i.e. every 1 s or across 30 s. The average classification accuracy was found to be slightly lower in the beginning of each 5 min sequence (see Figure 2c), both when training a model with less (500 training subjects) and more (75% of total training subjects). Interestingly, when training with less data, we also observed a lower accuracy in the beginning and end of each 30 s segment relative to the accuracy in the middle section, which was not the case when training with more data.

### Choice of cohort impacts classification performance on test set

In Figure 3 we show how training on different cohorts yield differing results in subsequent testing performance, here expressed in heatmaps as both overall accuracy (Figure 3a), and Cohen's kappa (Figure 3b) averaged across all $N = 1{,}584$ subject PSGs in the test partition. The first two columns show the performance on the cohort on the x-axis, when training on the specific cohort on the y-axis. Since the training subset in ISRUC is small compared to the other cohorts, we trained the model in the left-most column with weight decay of $10^{-4}$ to compensate for the risk of overfitting, however, by comparing the left and middle columns, we did not observe any specific gain in classification performance by doing so. The right-most column shows the test performance for each cohort, when excluding that cohort from training. We observe a significant spread in classification accuracy across the different cohorts with prediction on ISRUC being poorest, while prediction on MrOS data being best. Further details can be found in Table S2.

**More data is good, diverse data is better**

We observed a general increase in classification performance both in terms of overall accuracy and Cohen's kappa, when including more data in the model training phase in both the mixed- and single-cohort setting (Figure 4a, Table S3). Classification performance was consistently lower in the single-cohort setting compared to the corresponding mixed-cohort setting. Interestingly, we found that training a model with just 0.25% of mixed-cohort training data still achieved an acceptable accuracy comparable to training a model with only SHHS data, while using all available training data increased that performance by almost 10 percentage points. Furthermore, we observed that the model trained with 100% of the training partition reached a state-of-the-art level of performance with an overall accuracy of $0.869 \pm 0.064$ ($95\%$ CI: $[0.865 - 0.872]$) and Cohen's kappa of $0.799 \pm 0.098$ ($95\%$ CI: $[0.794 - 0.804]$) (Table S3). The model furthermore performs well with respect to classifying individual sleep stages as shown in the confusion matrix in Figure 4b. However, the model still has difficulties classifying and distinguishing between certain sleep stages, especially between N2, N1, and N3; and W, N2, and N1.

**Increasing the number of data sources improves classification performance**

On average, we saw an increase in overall accuracy, when increasing the number of cohorts from 2 to 4 using 500 PSGs in each configuration, see Figure 5 and Table S4. Specifically, we found that the average overall accuracy increased from $0.788 \pm 0.102$ ($95\%$ CI: $[0.787 - 0.790]$) in the 2-cohort configuration to $0.808 \pm 0.092$ ($95\%$ CI: $[0.807 - 0.810]$) and $0.821 \pm 0.085$ ($95\%$ CI: $[0.819 - 0.823]$) in the 4-cohort configuration.

**DISCUSSION**

In this work, we present an end-to-end deep learning-based model for fully automatic micro- and macro-sleep stage classification. Using all of the available data sources for training our model, we reached an overall accuracy on test partition of $0.869 \pm 0.064$ (95% CI: $[0.865 - 0.872]$), and a Cohen's kappa of $0.799 \pm 0.098$ (95% CI: $[0.794 - 0.804]$), which is in the very high end of the substantial agreement category for observer agreement[51]. We found that individual cohorts exhibit major differences in overall accuracy and Cohen's kappa when subjected to both training and testing conditions and specifically, we found that average performance on the test partition in the LOCI configurations varied significantly from $0.676 \pm 0.124$ (95% CI: $[0.670 - 0.682]$) when training on ISRUC, to $0.837 \pm 0.084$ (95% CI: $[0.833 - 0.841]$) when training on SHHS. Each individual cohort also showed large deviations in predictive performance when tested on the other cohorts. For example, when conditioned on SHHS data, the lowest average accuracy was 0.721 on SSC test data compared to the highest at 0.872 on SHHS test data, while conditioning on SSC training data, the lowest average accuracy was 0.704 on ISRUC test data compared to 0.824 on WSC test data. Classification performance was generally higher on the test set when using the LOCO configuration, except for SHHS (higher in LOCI) and SSC (no difference). We also found that having data from multiple sources always resulted in better-performing models compared to training on single cohorts. Increasing the number of data sources increased classification performance, although this was non-significant. In the design of the model, we observed that model performance was enhanced by the addition of the recurrent module (bGRU), a phenomenon likely reflecting the fact that sleep stage scoring at a specific time in one subject can be influenced by signal content (frequency, amplitude, presence of micro-events) at later time steps. However, the complexity of the module given by the number of hidden units did not affect performance. In all our experiments, we also evaluated the performance of the model every 1 s compared to the performance evaluated every 30 s and found them to be similar, which indicates the model is stable in classification in periods corresponding to an epoch of data.

Only a handful of studies have previously reported results when using multiple cohorts[21–23]. Some authors have reported a drop from 81.9% to 77.7% when training on the Massachusetts General Hospital cohort (MGH) and testing on MGH and SHHS, respectively[22], while others have shown significant drops from 89.8% to 81.4% and 72.1% on two separate hold-out sets from Singapore and USA[23]. We also observed similar trends in our LOCI and LOCO experiments, where excluding the training subset of a cohort from the training partition resulted in a significant drop in performance on the respective test subset from that cohort. A benefit of our LOCI and LOCO experiments is the possibility for direct benchmarking against previous publications using specific cohorts in their experiments. For example, we obtain an accuracy of 0.805 in the LOCO-SHHS training-testing case compared to 0.777 previously reported by Biswal et al.[22], both of which reflect classification performance when SHHS had not been used for training; and an accuracy of 0.865 in the LOCI-WSC case compared to 0.841 reported by Olesen et al. [24], where both have been using a subset of WSC for training the model. Interestingly, we obtained the same level of performance on the SHHS data in our LOCI experiment as reported by Sors et al. (87% accuracy, 81% Cohen's kappa) even though they only used single-EEG for their experiments[52]. Other works that have investigated single- vs. multi-channel models for automatic sleep stage classification have found that models generally benefit from having more channels available for training[16,18,22]. It may be that some cohorts share different characteristics that makes them more suitable for single- or multi-channel models, but this is speculative and would need to be verified in subsequent studies.

Our study is not without limitations. We only optimized our network architecture with respect to the temporal processing module and therefore cannot assess what impact different design choices for the other modules would have had on final performance. For example, the EMG signal has different statistical properties and spectral content, and separate, parallel architectures for EMG and EEG/EOG feature extraction may be warranted, as proposed by others[16,21]. Other studies have however shown equal performance in large cohorts using a similar channel mixing approach as proposed here[24]. Another limitation is found in our training runs, as we did not consider balancing our data with respect to the proportion of sleep stages, which may or may not have had impact on overall performance. It is well established that there is significant variation in scoring and validation of N1/REM and N2/N3[3,5,7], which challenges the training for any classification algorithm. Some researchers have experimented balancing the cost of misclassifying sleep stages by weighting them by their inverse frequency of occurrence and found no significant improvement[24,52], while others have experimented with balancing the sleep stage frequencies in each batch of data input to the neural network model[16], but more rigorous research in resampling or over/under-sampling techniques is warranted in this regard. We ultimately decided against experimenting with balancing our sleep stages in each batch, as we prioritized flexibility with regards to the length of input sequences fed to the network. All our models ran through at least 50 epochs of training (passes through the training partition), which might have induced a bias in the configurations with larger cohorts. For example, one pass through the training partition in the LOCI-ISRUC case corresponds to much less data than one pass through the LOCI-SHHS case. However, since we selected the best performing model based on Cohen's kappa across all 50 epochs, we have allowed for more effective training in cases with less available training data. We observed that models using less data in the training partition generally had to run for longer time (i.e. more epochs) before converging.

In future studies on automatic sleep stage classification algorithms, we strongly recommend researchers to test and report results on not just hold-out test partitions, but also on cohorts completely unseen by the model both during training and testing/validation. Our experiments indicate that even though good performance can be achieved on hold-out data using a single cohort, this does not necessarily translate into good generalization performance. Such approach requires availability of many publicly available, high-quality, well-documented databases with easily accessible PSG data, associated annotations and related patient information. In this regard, websites such as the NSRR, which contains several large databases with clinical data as well as PSG and annotation data in a standardized format[31,32], are an invaluable resource for researchers. We also propose that the sleep science community establishes a common reference dataset on which researchers in machine learning can benchmark their models, similar to what the computer vision and general machine learning community has done with the ImageNet Large Scale Visual Recognition Challenge (ILSVRC)[53], an annual competition in which researchers submit their models to test in various competitions.

In summary, we have developed an automatic sleep stage classification algorithm based on deep learning, that can accurately classify sleep stages at a flexible resolution with a state-of-the-art classification performance of 87% accuracy on a test set of 1,584 PSGs. We trained and tested our model using five cohorts with varying numbers of PSGs covering multiple phenotypes with specific focus on how well cohorts can generalize to each other. We found that different cohorts generalize very differently both in intra- and inter-cohort settings (LOCI vs. LOCO experiments). Furthermore, we also found that having more data sources significantly improve classification performance and generalizability to the extent that even just a small number of training PSGs can reach high classification performance by including many different sources. To our knowledge, this is one of the largest, if not the largest, study on automatic sleep stage classification in terms of PSG volume, diversity, and performance.

**ACKNOWLEDGMENTS**

Some of the computing for this project was performed on the Sherlock cluster. We would like to thank Stanford University and the Stanford Research Computing Center for providing computational resources and support that contributed to these research results.

The authors would also like to thank the National Sleep Research Resource (https://www.sleepdata.org) team for their work in collecting, organizing and making available some of the PSG data used in this study.

The authors would also like to thank Julie Anja Engelhard Christensen, PhD, for her editorial work on this paper.

**DISCLOSURE STATEMENT**

A. N. Olesen has received funding from The Klarman Family Foundation; Technical University of Denmark; University of Copenhagen; Reinholdt W. Jorck og Hustrus Fonden; Otto Mønsteds Fond; Stibofonden; Knud Højgaards Fond; Augustinus Fond; and Vera og Carl Johan Michaelsens Fond. E. Mignot has received partial funding by the Klarman Family Foundation, has received research support from Jazz Pharmaceuticals, is a consultant for Rhythm (a sleep consumer product company) and Alairion (a sleep apnea pharmacology company), and is on the speakers' bureau for Vox Media.

## REFERENCES

1. Chiao W, Durr ML. Trends in sleep studies performed for Medicare beneficiaries. *Laryngoscope*. 2017;127(12):2891-2896. doi:10.1002/lary.26736
2. Berry RB, Albertario CL, Harding SM, et al. *The AASM Manual for the Scoring of Sleep and Associated Events: Rules, Terminology and Technical Specifications.* 2.5. Darien, Il: American Academy of Sleep Medicine; 2018.
3. Younes M, Raneri J, Hanly P. Staging sleep in polysomnograms: Analysis of inter-scorer variability. *J Clin Sleep Med*. 2016;12(6):885-894. doi:10.5664/jcsm.5894
4. Younes M. The case for using digital EEG analysis in clinical sleep medicine. *Sleep Sci Pract*. 2017;1(2). doi:10.1186/s41606-016-0005-0
5. Younes M, Kuna ST, Pack AI, et al. Reliability of the American Academy of Sleep Medicine Rules for Assessing Sleep Depth in Clinical Practice. *J Clin Sleep Med*. 2018;14(02):205-213. doi:10.5664/jcsm.6934
6. Rosenberg RS, Van Hout S. The American Academy of Sleep Medicine Inter-scorer Reliability Program: Sleep Stage Scoring. *J Clin Sleep Med*. 2013;9(1):81-87. doi:10.5664/jcsm.2350
7. Norman RG, Pal I, Stewart C, Walsleben JA, Rapoport DM. Interobserver Agreement Among Sleep Scorers From Different Centers in a Large Dataset. *Sleep*. 2000;23(7):1-8. doi:10.1093/sleep/23.7.1e
8. Fiorillo L, Puiatti A, Papandrea M, et al. Automated sleep scoring: A review of the latest approaches. *Sleep Med Rev*. 2019;48:101204. doi:10.1016/j.smrv.2019.07.007
9. Goldberger AL, Amaral LAN, Glass L, et al. PhysioBank, PhysioToolkit, and PhysioNet. *Circulation*. 2000;101(23):e215-e220. doi:10.1161/01.CIR.101.23.e215
10. Kemp B, Zwinderman AH, Tuk B, Kamphuisen HAC, Oberyé JJL. Analysis of a sleep-dependent neuronal feedback loop: The slow-wave microcontinuity of the EEG. *IEEE Trans Biomed Eng*. 2000;47(9):1185-1194. doi:10.1109/10.867928
11. Vilamala A, Madsen KH, Hansen LK. Deep convolutional neural networks for interpretable analysis of EEG sleep stage scoring. In: *2017 IEEE 27th International Workshop on Machine Learning for Signal Processing (MLSP)*. Tokyo, Japan: IEEE; 2017:1-6. doi:10.1109/MLSP.2017.8168133
12. Phan H, Andreotti F, Cooray N, Chen OY, Vos M De. Automatic Sleep Stage Classification Using Single-Channel EEG: Learning Sequential Features with Attention-Based Recurrent Neural Networks. In: *2018 40th Annual International Conference of the IEEE Engineering in Medicine and Biology Society (EMBC)*. IEEE; 2018:1452-1455. doi:10.1109/EMBC.2018.8512480
13. Supratak A, Dong H, Wu C, Guo Y. DeepSleepNet: A Model for Automatic Sleep Stage Scoring Based on Raw Single-Channel EEG. *IEEE Trans Neural Syst Rehabil Eng*. 2017;25(11):1998-2008. doi:10.1109/TNSRE.2017.2721116
14. Şen B, Peker M, Çavuşoğlu A, Çelebi F V. A comparative study on classification of sleep stage based on EEG signals using feature selection and classification algorithms. *J Med Syst*. 2014;38(3). doi:10.1007/s10916-014-0018-0
15. O'Reilly C, Gosselin N, Carrier J, Nielsen T. Montreal archive of sleep studies: An open-access resource for instrument benchmarking and exploratory research. *J Sleep Res*. 2014;23(6):628-635. doi:10.1111/jsr.12169
16. Chambon S, Galtier MN, Arnal PJ, Wainrib G, Gramfort A. A Deep Learning Architecture for Temporal Sleep Stage Classification Using Multivariate and Multimodal Time Series. *IEEE Trans Neural Syst Rehabil Eng*. 2018;26(4):758-769. doi:10.1109/TNSRE.2018.2813138
17. Andreotti F, Phan H, Cooray N, Lo C, Hu MTM, De Vos M. Multichannel Sleep Stage Classification and Transfer Learning using Convolutional Neural Networks. In: *2018 40th Annual International Conference of the IEEE Engineering in Medicine and Biology Society (EMBC)*. IEEE; 2018:171-174. doi:10.1109/EMBC.2018.8512214
18. Phan H, Andreotti F, Cooray N, Chen OY, De Vos M. Joint Classification and Prediction CNN Framework for Automatic Sleep Stage Classification. *IEEE Trans Biomed Eng*. 2019;66(5):1285-1296. doi:10.1109/TBME.2018.2872652

19. Phan H, Andreotti F, Cooray N, Chen OY, De Vos M. SeqSleepNet: End-to-End Hierarchical Recurrent Neural Network for Sequence-to-Sequence Automatic Sleep Staging. *IEEE Trans Neural Syst Rehabil Eng*. 2019;27(3):400-410. doi:10.1109/TNSRE.2019.2896659

20. LeCun Y, Bengio Y, Hinton G. Deep learning. *Nature*. 2015;521(7553):436-444. doi:10.1038/nature14539

21. Stephansen JB, Olesen AN, Olsen M, et al. Neural network analysis of sleep stages enables efficient diagnosis of narcolepsy. *Nat Commun*. 2018;9(1):5229. doi:10.1038/s41467-018-07229-3

22. Biswal S, Sun H, Goparaju B, Westover MB, Sun J, Bianchi MT. Expert-level sleep scoring with deep neural networks. *J Am Med Informatics Assoc*. 2018;25(12):1643-1650. doi:10.1093/jamia/ocy131

23. Patanaik A, Ong JL, Gooley JJ, Ancoli-Israel S, Chee MWL. An end-to-end framework for real-time automatic sleep stage classification. *Sleep*. 2018;41(5):1-11. doi:10.1093/sleep/zsy041

24. Olesen AN, Jennum P, Peppard P, Mignot E, Sorensen HBD. Deep residual networks for automatic sleep stage classification of raw polysomnographic waveforms. In: *2018 40th Annual International Conference of the IEEE Engineering in Medicine and Biology Society (EMBC)*. IEEE; 2018:1-4. doi:10.1109/EMBC.2018.8513080

25. Biswal S, Kulas J, Sun H, et al. SLEEPNET: Automated Sleep Staging System via Deep Learning. July 2017:1-17. http://arxiv.org/abs/1707.08262.

26. Khalighi S, Sousa T, Santos JM, Nunes U. ISRUC-Sleep: A comprehensive public dataset for sleep researchers. *Comput Methods Programs Biomed*. 2016;124:180-192. doi:10.1016/j.cmpb.2015.10.013

27. Blank JB, Cawthon PM, Carrion-Petersen M Lou, et al. Overview of recruitment for the osteoporotic fractures in men study (MrOS). *Contemp Clin Trials*. 2005;26(5):557-568. doi:10.1016/j.cct.2005.05.005

28. Orwoll E, Blank JB, Barrett-Connor E, et al. Design and baseline characteristics of the osteoporotic fractures in men (MrOS) study — A large observational study of the determinants of fracture in older men. *Contemp Clin Trials*. 2005;26(5):569-585. doi:10.1016/j.cct.2005.05.006

29. Blackwell T, Yaffe K, Ancoli-Israel S, et al. Associations Between Sleep Architecture and Sleep-Disordered Breathing and Cognition in Older Community-Dwelling Men: The Osteoporotic Fractures in Men Sleep Study. *J Am Geriatr Soc*. 2011;59(12):2217-2225. doi:10.1111/j.1532-5415.2011.03731.x

30. Rechtschaffen A, Kales A, eds. *A Manual of Standardized Terminology, Techniques and Scoring System for Sleep Stages of Human Subjects*. Washington, DC: National Institute of Health; 1968.

31. Dean DA, Goldberger AL, Mueller R, et al. Scaling Up Scientific Discovery in Sleep Medicine: The National Sleep Research Resource. *Sleep*. 2016;39(5):1151-1164. doi:10.5665/sleep.5774

32. Zhang G-Q, Cui L, Mueller R, et al. The National Sleep Research Resource: towards a sleep data commons. *J Am Med Informatics Assoc*. 2018;0(June):1-8. doi:10.1093/jamia/ocy064

33. Redline S, Sanders MH, Lind BK, et al. Methods for obtaining and analyzing unattended polysomnography data for a multicenter study. Sleep Heart Health Research Group. *Sleep*. 1998;21(7):759-767. http://www.ncbi.nlm.nih.gov/pubmed/11300121.

34. Quan SF, Howard B V, Iber C, et al. The Sleep Heart Health Study: design, rationale, and methods. *Sleep*. 1997;20(12):1077-1085. http://www.ncbi.nlm.nih.gov/pubmed/9493915.

35. Young T, Finn L, Peppard PE, et al. Sleep Disordered Breathing and Mortality: Eighteen-Year Follow-up of the Wisconsin Sleep Cohort. *Sleep*. 2008;31(8):291-292. doi:10.5665/sleep/31.8.1071

36. Young T, Palta M, Dempsey J, Skatrud J, Weber S, Badr S. The Occurrence of Sleep-Disordered Breathing among Middle-Aged Adults. *N Engl J Med*. 1993;328(17):1230-1235. doi:10.1056/NEJM199304293281704

37. Andlauer O, Moore H, Jouhier L, et al. Nocturnal Rapid Eye Movement Sleep Latency for Identifying Patients With Narcolepsy/Hypocretin

Deficiency. *JAMA Neurol*. 2013;70(7):891. doi:10.1001/jamaneurol.2013.1589

38. Moore H, Leary E, Lee S-Y, et al. Design and Validation of a Periodic Leg Movement Detector. *PLoS One*. 2014;9(12):e114565. doi:10.1371/journal.pone.0114565

39. Chambon S, Thorey V, Arnal PJ, Mignot E, Gramfort A, Neurospin CEA. A deep learning architecture to detect events in EEG signals during sleep. In: *2018 IEEE 28th International Workshop on Machine Learning for Signal Processing (MLSP)*. IEEE; 2018:1-6.

40. Chambon S, Thorey V, Arnal PJ, Mignot E, Gramfort A. DOSED: A deep learning approach to detect multiple sleep micro-events in EEG signal. *J Neurosci Methods*. 2019;321:64-78. doi:10.1016/j.jneumeth.2019.03.017

41. Olesen AN, Chambon S, Thorey V, Jennum P, Mignot E, Sorensen HBD. Towards a Flexible Deep Learning Method for Automatic Detection of Clinically Relevant Multi-Modal Events in the Polysomnogram. In: *2019 41st Annual International Conference of the IEEE Engineering in Medicine and Biology Society (EMBC)*. IEEE; 2019:556-561. doi:10.1109/EMBC.2019.8856570

42. Ioffe S, Szegedy C. Batch Normalization: Accelerating Deep Network Training by Reducing Internal Covariate Shift. In: *Proceedings of the 32nd International Conference on Machine Learning*. Vol 37. Lille, France: JMLR: W&CP; 2015. doi:10.1007/s13398-014-0173-7.2

43. He K, Zhang X, Ren S, Sun J. Identity Mappings in Deep Residual Networks. In: *Computer Vision -- ECCV 2016*. Vol abs/1603.0. ; 2016:630-645. doi:10.1007/978-3-319-46493-0_38

44. He K, Zhang X, Ren S, Sun J. Deep Residual Learning for Image Recognition. In: *IEEE Conference on Computer Vision and Pattern Recognition (CVPR)*. ; 2015:770-778. doi:10.1109/CVPR.2016.90

45. Cho K, van Merrienboer B, Bahdanau D, Bengio Y. On the Properties of Neural Machine Translation: Encoder–Decoder Approaches. In: *Proceedings of SSST-8, Eighth Workshop on Syntax, Semantics and Structure in Statistical Translation*. Stroudsburg, PA, USA: Association for Computational Linguistics; 2014:103-111. doi:10.3115/v1/W14-4012

46. Kingma DP, Ba J. Adam: A Method for Stochastic Optimization. In: *3rd International Conference on Learning Representations (ICLR)*. San Diego, CA; 2015:1-15. http://arxiv.org/abs/1412.6980.

47. Glorot X, Bengio Y. Understanding the difficulty of training deep feedforward neural networks. *Proc Thirteen Int Conf Artif Intell Stat PMLR*. 2010;9:249-256.

48. Paszke A, Gross S, Chintala S, et al. Automatic differentiation in PyTorch. In: *31st Conference on Neural Information Processing Systems (NIPS)*. Long Beach, CA, USA; 2017.

49. Pedregosa F, Varoquaux G, Gramfort A, et al. Scikit-learn: Machine Learning in Python. *J Mach Learn Res*. 2011;12:2825-2830.

50. Brink-Kjaer A, Olesen AN, Peppard PE, et al. *Automatic Detection of Cortical Arousals in Sleep and Their Contribution to Daytime Sleepiness*.; 2019. http://arxiv.org/abs/1906.01700.

51. Landis JR, Koch GG. The Measurement of Observer Agreement for Categorical Data. *Biometrics*. 1977;33:159-174.

52. Sors A, Bonnet S, Mirek S, Vercueil L, Payen JF. A convolutional neural network for sleep stage scoring from raw single-channel EEG. *Biomed Signal Process Control*. 2018;42:107-114. doi:10.1016/j.bspc.2017.12.001

53. Russakovsky O, Deng J, Su H, et al. ImageNet Large Scale Visual Recognition Challenge. *Int J Comput Vis*. 2015:211-252. doi:10.1007/s11263-015-0816-y

## FIGURE CAPTIONS LIST

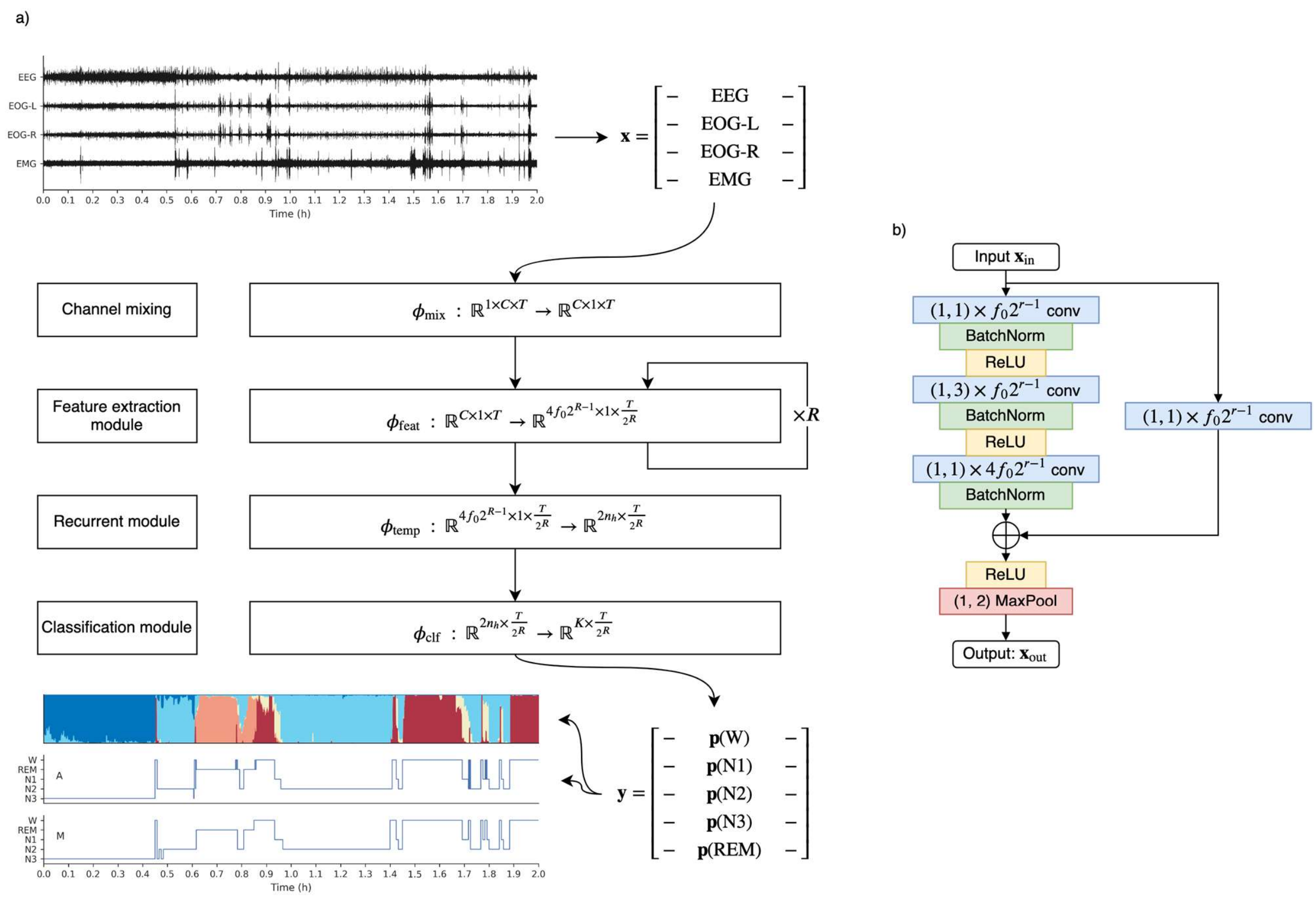

Figure 1: Model overview. a) The input is a sequence of data **x** containing raw signal data from EEG, EOG-L/R, and EMG channels, which is supplied to the network modules in sequence. The feature extraction module consists of $R$ repeated blocks of residual units, see panel to the right. The output of the model is a matrix **y** containing class probabilities for each sleep stage for each time step, which can be visualized either directly as a hypnodensity, or by $\arg\max \mathbf{y}$ as a hypnogram. The "A" and "M" labels in the hypnogram plots corresponds to automatic and manual hypnograms. b) Schematic of a single residual block in the feature extraction module. Convolutional layers are described by the kernel size × number of filters using a stride value of 1. Shortcut uses 1x1 convolutions with added zero-padding to maintain temporal dimension. Conv, convolutional layer; BatchNorm, batch normalization; ReLU, rectified linear unit; $f_0$, base number of filters ($f_0 = 4$).

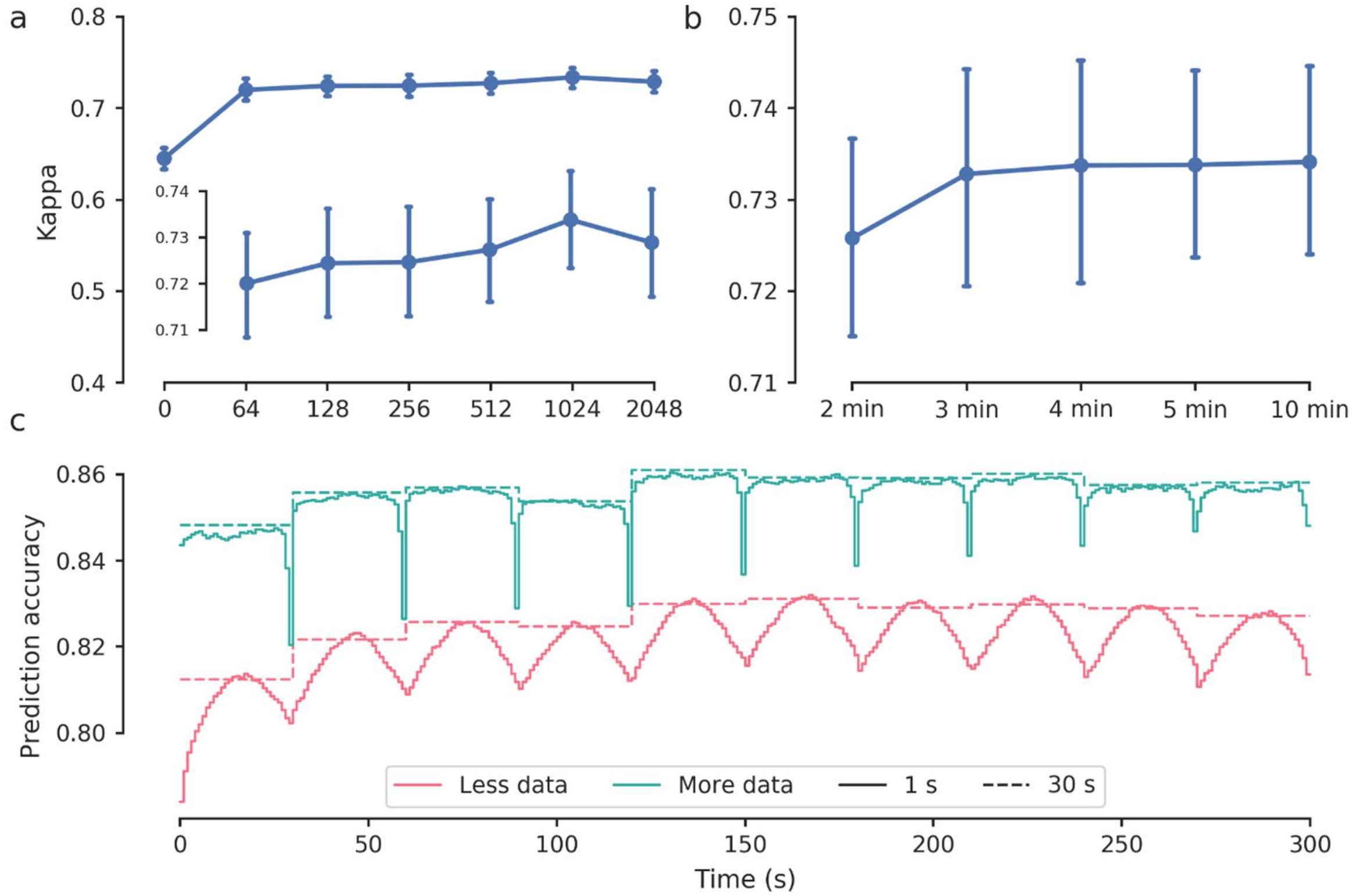

Figure 2: Temporal context changes model performance. a) Cohen's kappa as a function of the number of hidden units in the recurrent block. Inset shows zoom of Cohen's kappa for non-zero hidden unit values. b) Cohen's kappa as a function of sequence length. c) Prediction accuracy averaged across all 5-minute sequences in the test partition with a small and large training partition. Full lines are predictions evaluated every 1 s, while dashed lines show predictions averaged every 30 s. Values are shown for panels a), b) as mean with 95% confidence intervals.

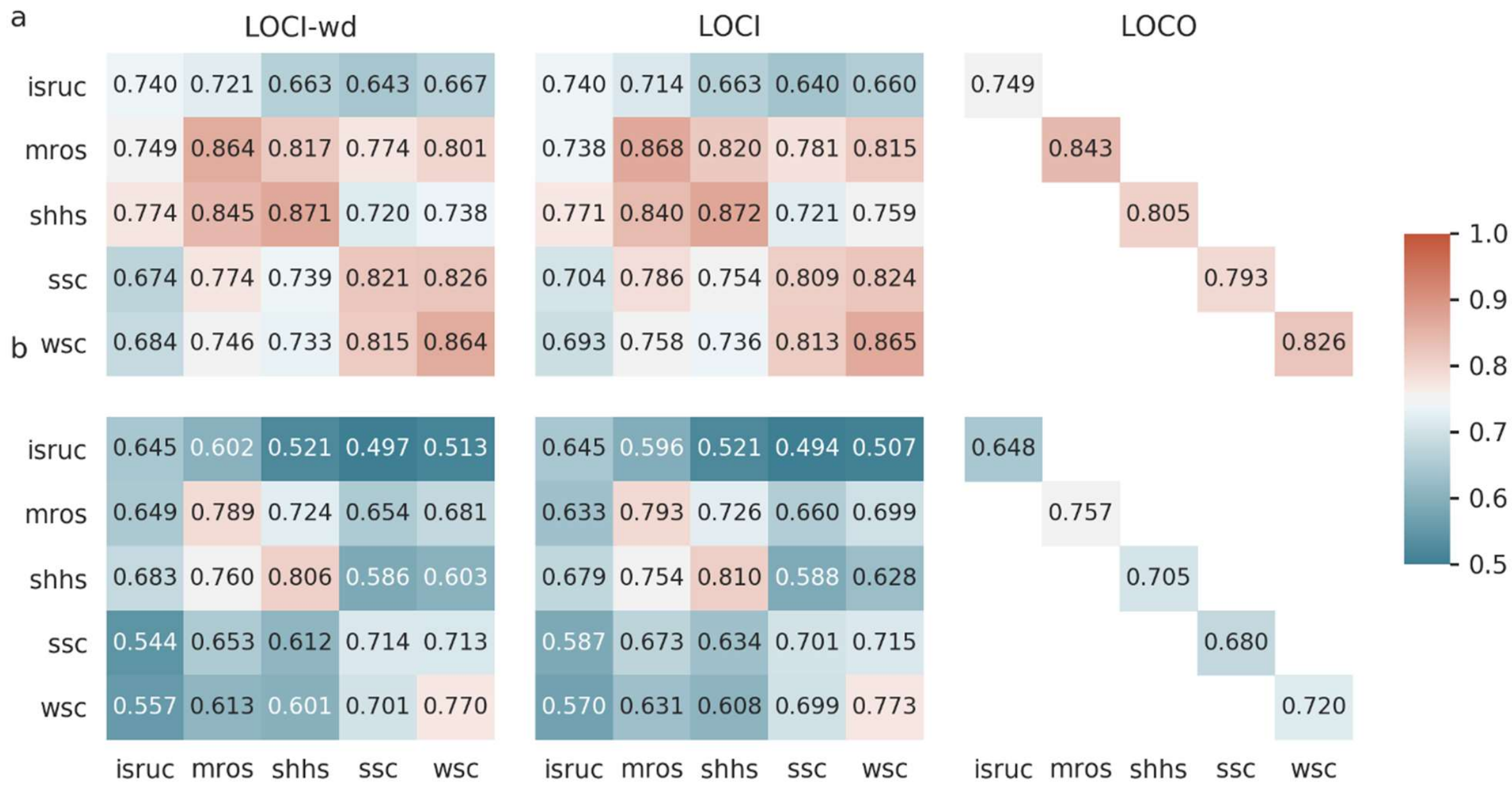

Figure 3: Individual cohorts influence classification performance on test partition ($N = 1{,}584$). As an example, training on MrOS in a LOCI configuration, the performance on the test subset of WSC is 0.815. The diagonals in all three configurations shows the performance for the same subjects in the test subsets in the respective cohorts making possible direct comparisons between LOCI and LOCO. For aggregated metrics and more summary statistics, please see ***Error! Reference source not found.***. LOCI, leave-one-cohort-in; LOCI-wd, LOCI with weight decay; LOCO, leave-one-cohort-out; ISRUC, Institute of Systems and Robotics, University of Coimbra Sleep Cohort; MrOS, MrOS Sleep Study; SHHS, Sleep Heart Health Study; SSC, Stanford Sleep Cohort; WSC, Wisconsin Sleep Cohort.

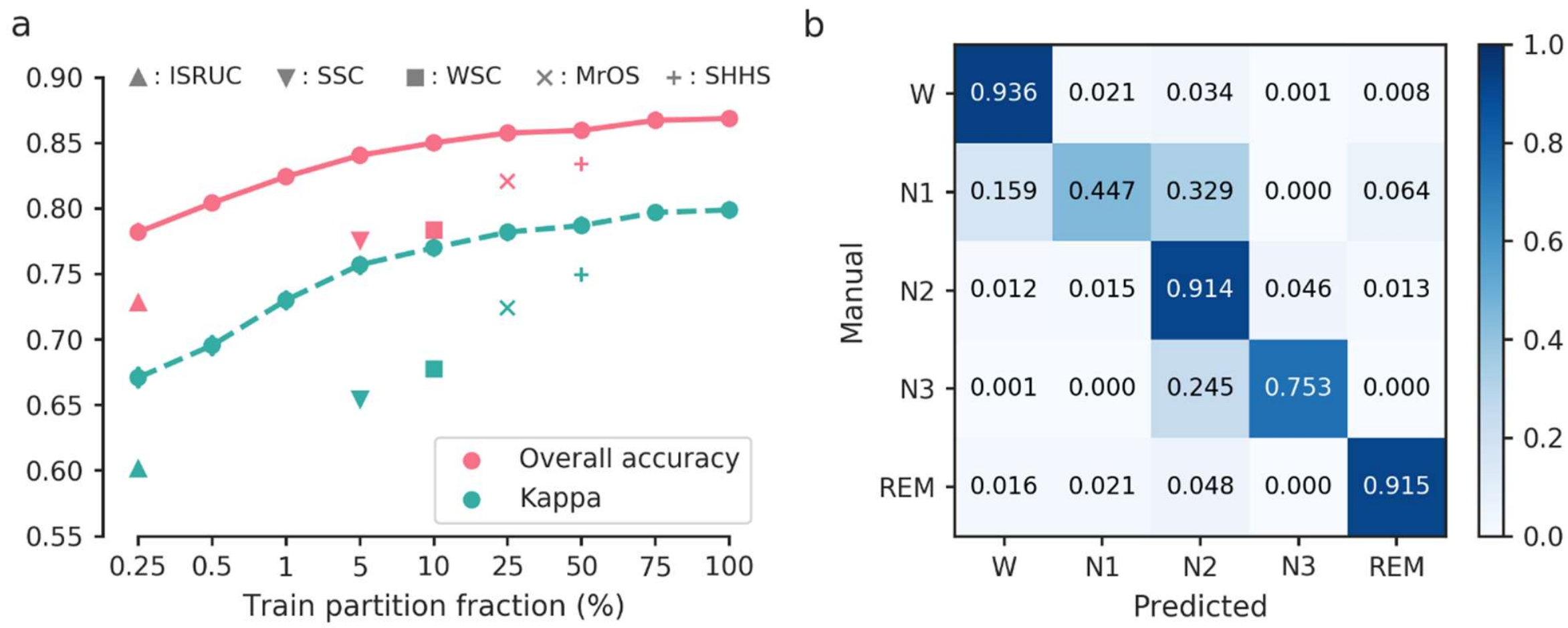

Figure 4: Training on mixed data increased predictive performance compared to individual cohorts of similar size. a) There is a gain in predictive performance by mixing data from various sources consistent across the size of the training dataset. b) Confusion matrix for a model trained on 100% of the available training partition data. The model shows excellent performance overall, with most misclassification happening between W and N1, and N1, N2, and N3. This is somewhat consistent with clinical experience, since N1 is a transition stage between wake and the deeper stages of sleep with much frequency content overlap with both W and N2.

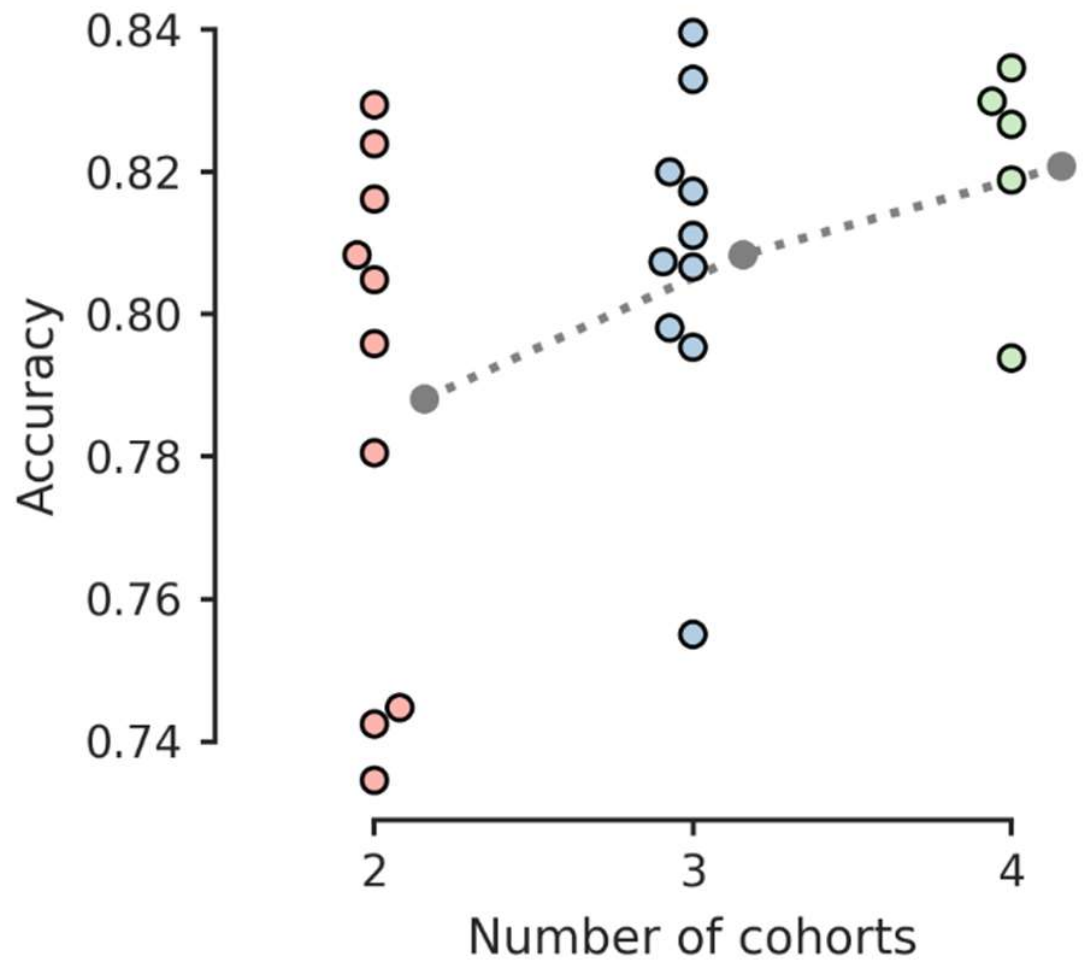

Figure 5: Number of cohorts in training partition increases model performance. Each datapoint is shown as the overall accuracy aggregated across all subjects for a specific training configuration. For example, the bottom dot in column 2 (3 cohort configuration) shows the performance on the test set (overall accuracy $0.755 \pm 0.109$, $95\%$ CI: $[0.750 - 0.760]$), when training with 500 PSGs randomly and evenly drawn from the Stanford Sleep Cohort, the Institute of Systems and Robotics, University of Coimbra Sleep Cohort, and the Wisconsin Sleep Cohort. Notice the scale on the y-axis.

**TABLES**

Table 1: Cohort demographics.

| | **ISRUC** | **MrOS** | **SHHS** | **SSC** | **WSC** | ***p*-value** |
|---|---|---|---|---|---|---|
| N (female) | 126 (50) | 3932 (0) | 8444 (4458) | 767 (319) | 2401 (1103) | 0 |
| Age, years | 49.8 ± 15.9 [20.0-85.0] | 77.6 ± 5.6 [67.0-90.0] | 64.5 ± 11.2 [39.0-90.0] | 45.7 ± 14.5 [13.0-104.8] | 59.7 ± 8.4 [37.2-82.3] | 0 |
| BMI, kg/m2 | - | 27.1 ± 3.8 [16.0-47.0] | 28.2 ± 5.1 [18.0-50.0] | 27.2 ± 6.5 [9.8-78.7] | 31.6 ± 7.2 [17.5-70.6] | 1.03e-171 |
| TST, min | 350.0 ± 67.3 [87.5-479.0] | 352.1 ± 71.9 [39.0-626.0] | 374.1 ± 69.4 [68.0-605.0] | 361.0 ± 83.5 [0.0-661.0] | 364.1 ± 63.6 [19.5-575.0] | 4.07e-38 |
| SL, min | 17.7 ± 20.5 [0.0-144.5] | 24.7 ± 26.9 [1.0-402.0] | 24.2 ± 25.7 [0.0-349.0] | 93.5 ± 58.9 [0.5-404.0] | 33.2 ± 21.4 [0.5-333.0] | 0 |
| REML, min | 125.6 ± 61.4 [7.0-323.0] | 104.8 ± 75.1 [0.0-590.0] | 91.7 ± 58.8 [0.0-471.0] | 140.9 ± 88.0 [0.0-464.0] | 128.3 ± 76.0 [3.5-514.0] | 2.81e-173 |
| WASO, min | 76.2 ± 49.8 [7.5-251.0] | 117.5 ± 67.6 [4.0-487.0] | 80.2 ± 54.7 [2.0-378.0] | 79.5 ± 55.0 [3.5-367.0] | 73.6 ± 45.9 [3.0-325.0] | 4.74e-233 |
| SE, % | 78.8 ± 14.1 [19.5-98.3] | 75.5 ± 12.4 [12.0-99.0] | 80.5 ± 11.0 [11.3-99.0] | 77.4 ± 14.8 [0.0-98.0] | 77.1 ± 11.2 [4.1-95.6] | 4.23e-117 |
| N1, % | 13.3 ± 5.8 [1.8-33.1] | 8.3 ± 6.4 [0.0-70.0] | 5.5 ± 4.0 [0.0-39.1] | 11.7 ± 10.2 [0.0-92.0] | 10.8 ± 6.9 [1.0-88.4] | 0 |
| N2, % | 31.9 ± 10.3 [4.4-89.3] | 62.5 ± 10.0 [21.0-95.0] | 56.9 ± 11.5 [10.9-100.0] | 62.8 ± 24.9 [0.0-636.0] | 66.0 ± 9.4 [9.1-93.3] | 0 |
| N3, % | 19.6 ± 8.0 [0.0-41.1] | 36.0 ± 31.8 [0.0-259.0] | 17.5 ± 11.6 [0.0-70.1] | 9.0 ± 9.3 [0.0-73.0] | 7.2 ± 7.8 [0.0-47.5] | 0 |
| REM, % | 13.3 ± 6.3 [0.0-37.8] | 19.3 ± 6.8 [0.0-44.0] | 20.1 ± 6.3 [0.0-48.0] | 16.3 ± 7.2 [0.0-40.0] | 16.0 ± 6.2 [0.0-38.2] | 1.12e-203 |
| ArI, /h | 20.2 ± 10.0 [2.1-72.0] | 23.7 ± 12.1 [1.0-105.0] | 18.9 ± 10.5 [0.0-110.4] | 125.0 ± 124.2 [1.0-729.0] | - | 0 |
| AHI, /h | 13.1 ± 13.2 [0.0-82.2] | 13.7 ± 14.6 [0.0-89.0] | 18.1 ± 16.2 [0.0-161.8] | 13.5 ± 19.2 [0.0-98.6] | 7.0 ± 9.4 [0.0-72.6] | 0 |
| PLMI, /h | 8.0 ± 27.4 [0.0-292.8] | 35.7 ± 37.5 [0.0-233.0] | - | 7.0 ± 18.1 [0.0-139.9] | - | 1.22e-169 |

Cohort data represented as mean ± SD [range] unless noted. Arousal annotations were not available for WSC; PLMI was not available for SHHS and WSC; BMI was not available for ISRUC. N: number of subjects; TST: total sleep time; SL: sleep latency; REML: REM latency; WASO: wake after sleep onset; SE: sleep efficiency; ArI: arousal index; AHI: apnea/hypopnea index; PLMI: periodic leg movement index; ; ISRUC: Institute of Systems and Robotics, University of Coimbra Sleep Cohort; MrOS: The Osteoporotic Fractures in Men Sleep Study; SHHS: Sleep Heart Health Study; SSC: Stanford Sleep Cohort; WSC: Wisconsin Sleep Cohort.

Table 2: Overview of model architecture.

| Module | Type | # filters/units | Kernel size | Stride | Activation | Output size |
|---|---|---|---|---|---|---|
| $\mathbf{x}$ | Input | – | – | – | – | $1 \times C \times T$ |
| $\boldsymbol{\varphi}_{\text{mix}}$ | 2D convolution | $C$ | $(1, C)$ | 1 | – | $C \times 1 \times T$ |
| | Batch normalization | – | – | 1 | ReLU | $C \times 1 \times T$ |
| $\boldsymbol{\varphi}_{\text{feat}}^{(r)}$, $\boldsymbol{r} \in [\![\boldsymbol{R}]\!]$ | †Residual module | $f_0 2^{r-1}/f_0 2^{r-1}/4 f_0 2^{r-1}$ | $(1,1)/(1,3)/(1,1)$ | $(1,1)/(1,2)$ | ReLU | $f_0 2^r \times 1 \times T/2^r$ |
| $\boldsymbol{\varphi}_{\text{temp}}$ | Bidirectional GRU | $n_h$ | – | – | – | $2n_h \times T/2^R$ |
| $\boldsymbol{\varphi}_{\text{clf}}$ | 1D convolution | $K$ | $2n_h$ | 1 | Softmax | $K \times T/2^R$ |

Kernel sizes correspond to the first, second and third convolutional layer in each residual block. Stride counts correspond to the residual block and the subsequent maximum pooling operation. ReLU, rectified linear unit; GRU, gated recurrent unit; $C$, number of input channels; $T$, length of segment in samples; $f_0$, base number of filters in residual blocks; $R$, number of residual blocks; $n_h$, number of hidden units in GRU; $K$, number of sleep stage classes. †SeeFigure 1.

**SUPPLEMENTARY TABLES**

Table S1: Temporal context impact on model performance in validation partition ($n = 426$).

| | **Overall accuracy** | | | | **Cohen's kappa** | | | |
|---|---|---|---|---|---|---|---|---|
| | **Mean** | **SD** | **Median** | **95% CI, mean** | **Mean** | **SD** | **Median** | **95% CI, mean** |
| **Hidden units** | | | | | | | | |
| 0 | 0.779 | 0.083 | 0.794 | [0.771-0.787] | 0.645 | 0.126 | 0.660 | [0.633-0.657] |
| 64 | 0.818 | 0.079 | 0.837 | [0.810-0.825] | 0.720 | 0.120 | 0.745 | [0.709-0.731] |
| 128 | 0.821 | 0.080 | 0.841 | [0.813-0.829] | 0.724 | 0.121 | 0.745 | [0.713-0.736] |
| 256 | 0.820 | 0.082 | 0.843 | [0.812-0.828] | 0.725 | 0.124 | 0.751 | [0.713-0.736] |
| 512 | 0.822 | 0.079 | 0.841 | [0.815-0.830] | 0.727 | 0.119 | 0.752 | [0.716-0.739] |
| 1024 | 0.828 | 0.072 | 0.845 | [0.821-0.835] | 0.734 | 0.111 | 0.758 | [0.723-0.744] |
| 2048 | 0.823 | 0.080 | 0.843 | [0.816-0.831] | 0.729 | 0.122 | 0.757 | [0.717-0.740] |
| **Sequence length** | | | | | | | | |
| 2 min | 0.821 | 0.075 | 0.840 | [0.814-0.828] | 0.726 | 0.114 | 0.754 | [0.715-0.737] |
| 3 min | 0.826 | 0.080 | 0.845 | [0.818-0.833] | 0.733 | 0.123 | 0.762 | [0.721-0.744] |
| 4 min | 0.828 | 0.079 | 0.849 | [0.820-0.835] | 0.734 | 0.122 | 0.762 | [0.722-0.745] |
| 5 min | 0.828 | 0.072 | 0.845 | [0.821-0.835] | 0.734 | 0.111 | 0.758 | [0.723-0.744] |
| 10 min | 0.829 | 0.075 | 0.848 | [0.822-0.836] | 0.734 | 0.113 | 0.759 | [0.723-0.745] |
| **Window length** | | | | | | | | |
| 1 s | 0.824 | 0.074 | 0.843 | [0.817-0.831] | 0.728 | 0.113 | 0.752 | [0.717-0.738] |
| 3 s | 0.824 | 0.074 | 0.845 | [0.817-0.832] | 0.728 | 0.113 | 0.752 | [0.717-0.739] |
| 5 s | 0.825 | 0.074 | 0.843 | [0.818-0.832] | 0.728 | 0.113 | 0.752 | [0.717-0.739] |
| 10 s | 0.825 | 0.074 | 0.844 | [0.818-0.832] | 0.729 | 0.113 | 0.753 | [0.718-0.739] |
| 15 s | 0.826 | 0.074 | 0.845 | [0.818-0.833] | 0.729 | 0.113 | 0.755 | [0.719-0.740] |
| 30 s | 0.829 | 0.075 | 0.848 | [0.822-0.836] | 0.734 | 0.113 | 0.759 | [0.723-0.745] |

The **Hidden units** variable corresponds to varying the complexity in the recurrent module by increasing the number of hidden units. **Sequence length** indicate the length of the sequence of 30 epochs, while **Window length** correspond to varying the evaluation frequency.

Table S2: Performance characteristics for LOCI and LOCO training configurations.

| | *N* PSGs | Overall accuracy | | | | Cohen's kappa | | | |
|---|---|---|---|---|---|---|---|---|---|
| | | **Mean** | **SD** | **Median** | **95% CI, mean** | **Mean** | **SD** | **Median** | **95% CI, mean** |
| **LOCI-wd** | | | | | | | | | |
| ISRUC | 1584 | 0.679 | 0.123 | 0.701 | [0.673-0.685] | 0.542 | 0.169 | 0.574 | [0.533-0.550] |
| MrOS | 1584 | 0.821 | 0.077 | 0.835 | [0.817-0.825] | 0.727 | 0.114 | 0.745 | [0.721-0.733] |
| SHHS | 1584 | 0.834 | 0.088 | 0.858 | [0.830-0.839] | 0.750 | 0.132 | 0.786 | [0.744-0.757] |
| SSC | 1584 | 0.762 | 0.094 | 0.774 | [0.757-0.767] | 0.639 | 0.129 | 0.654 | [0.633-0.646] |
| WSC | 1584 | 0.758 | 0.105 | 0.773 | [0.753-0.764] | 0.633 | 0.145 | 0.653 | [0.626-0.640] |
| **LOCI** | | | | | | | | | |
| ISRUC | 1584 | 0.676 | 0.124 | 0.700 | [0.670-0.682] | 0.539 | 0.170 | 0.574 | [0.531-0.547] |
| MrOS | 1584 | 0.826 | 0.074 | 0.839 | [0.822-0.829] | 0.732 | 0.111 | 0.748 | [0.726-0.737] |
| SHHS‡ | 1584 | 0.837 | 0.084 | 0.858 | [0.833-0.841] | 0.754 | 0.127 | 0.786 | [0.748-0.761] |
| SSC | 1584 | 0.773 | 0.088 | 0.785 | [0.769-0.777] | 0.657 | 0.125 | 0.671 | [0.651-0.663] |
| WSC | 1584 | 0.763 | 0.101 | 0.776 | [0.758-0.768] | 0.641 | 0.140 | 0.659 | [0.635-0.648] |
| **LOCO** | | | | | | | | | |
| ISRUC† | 52 | 0.749 | 0.081 | 0.764 | [0.727-0.771] | 0.648 | 0.119 | 0.682 | [0.616-0.680] |
| | 126 | *0.757* | *0.071* | *0.766* | *[0.744-0.769]* | *0.661* | *0.101* | *0.682* | *[0.643-0.678]* |
| MrOS† | 371 | 0.843 | 0.066 | 0.851 | [0.836-0.849] | 0.757 | 0.104 | 0.776 | [0.746-0.767] |
| | 3932 | *0.841* | *0.069* | *0.854* | *[0.838-0.843]* | *0.752* | *0.107* | *0.775* | *[0.749-0.755]* |
| SHHS | 846 | 0.805 | 0.076 | 0.815 | [0.800-0.810] | 0.705 | 0.109 | 0.722 | [0.698-0.712] |
| | 8444 | *0.800* | *0.081* | *0.811* | *[0.798-0.801]* | *0.697* | *0.115* | *0.713* | *[0.694-0.699]* |
| SSC | 76 | 0.793 | 0.086 | 0.809 | [0.744-0.812] | 0.680 | 0.120 | 0.700 | [0.653-0.707] |
| | 766 | *0.798* | *0.086* | *0.815* | *[0.792-0.805]* | *0.690* | *0.123* | *0.711* | *[0.681-0.699]* |
| WSC† | 239 | 0.826 | 0.065 | 0.835 | [0.818-0.834] | 0.720 | 0.096 | 0.736 | [0.708-0.732] |
| | 2411 | *0.824* | *0.068* | *0.837* | *[0.821-0.827]* | *0.718* | *0.100* | *0.736* | *[0.714-0.722]* |

Metrics are aggregated across all subjects for each cohort in test partition ($N = 1{,}584$ PSGs). Statistics in italics correspond to evaluating performance on entire cohort. PSG: polysomnography; LOCI-wd: leave-one-cohort-in with weight decay; LOCO: leave-one-cohort-out; ISRUC: Institute of Systems and Robotics, University of Coimbra Sleep Cohort; MrOS: The Osteoporotic Fractures in Men Sleep Study; SHHS: Sleep Heart Health Study; SSC: Stanford Sleep Cohort; WSC: Wisconsin Sleep Cohort; †: significantly better than corresponding LOCI; ‡: significantly better than corresponding LOCO.

Table S3: Model performance of test partition with varying fractions of training data.

| | Overall accuracy | | | | Cohen's kappa | | | |
|---|---|---|---|---|---|---|---|---|
| | **Mean** | **SD** | **Median** | **95% CI, mean** | **Mean** | **SD** | **Median** | **95% CI, mean** |
| **Fraction (%)** | | | | | | | | |
| 0.25 | 0.782 | 0.097 | 0.801 | [0.777-0.787] | 0.671 | 0.141 | 0.696 | [0.664-0.678] |
| 0.50 | 0.804 | 0.086 | 0.824 | [0.800-0.808] | 0.696 | 0.131 | 0.724 | [0.689-0.702] |
| 1 | 0.824 | 0.079 | 0.840 | [0.820-0.828] | 0.730 | 0.118 | 0.753 | [0.724-0.736] |
| 5 | 0.841 | 0.074 | 0.856 | [0.837-0.844] | 0.757 | 0.113 | 0.780 | [0.751-0.763] |
| 10 | 0.850 | 0.069 | 0.864 | [0.847-0.853] | 0.770 | 0.108 | 0.791 | [0.765-0.775] |
| 25 | 0.858 | 0.066 | 0.873 | [0.854-0.861] | 0.782 | 0.102 | 0.804 | [0.777-0.787] |
| 50 | 0.860 | 0.063 | 0.874 | [0.856-0.863] | 0.787 | 0.097 | 0.809 | [0.782-0.792] |
| 75 | 0.867 | 0.062 | 0.882 | [0.864-0.870] | 0.797 | 0.096 | 0.818 | [0.792-0.802] |
| 100 | 0.869 | 0.064 | 0.883 | [0.865-0.872] | 0.799 | 0.098 | 0.820 | [0.794-0.804] |

Increasing the available training data increased performance on the test partition ($N = 1{,}584$) shown here as aggregated metrics across all subjects. No statistical difference was found by comparing confidence intervals (CI) between models trained with 75% and 100% of available training data, which indicates a saturation in training.

Table S4: Model performance on test partition ($N = 1{,}584$) with varying number of cohorts in training partition.

| **Training cohorts** | **Overall accuracy** | | | | **Kappa** | | | |
|---|---|---|---|---|---|---|---|---|
| | **Mean** | **SD** | **Median** | **95% CI, mean** | **Mean** | **SD** | **Median** | **95% CI, mean** |
| **2** | | | | | | | | |
| **Overall** | 0.788 | 0.102 | 0.811 | [0.787-0.790] | 0.683 | 0.143 | 0.710 | [0.681-0.685] |
| ISRUC-MrOS | 0.781 | 0.102 | 0.804 | [0.776-0.786] | 0.675 | 0.143 | 0.703 | [0.668-0.682] |
| ISRUC-SHHS | 0.808 | 0.097 | 0.835 | [0.804-0.813] | 0.717 | 0.142 | 0.756 | [0.710-0.724] |
| ISRUC-SSC | 0.735 | 0.103 | 0.753 | [0.729-0.740] | 0.613 | 0.140 | 0.638 | [0.606-0.620] |
| ISRUC-WSC | 0.745 | 0.107 | 0.758 | [0.740-0.750] | 0.628 | 0.140 | 0.642 | [0.621-0.635] |
| MrOS-SHHS | 0.829 | 0.081 | 0.849 | [0.825-0.833] | 0.740 | 0.124 | 0.769 | [0.734-0.746] |
| MrOS-SSC | 0.796 | 0.090 | 0.816 | [0.791-0.800] | 0.683 | 0.133 | 0.708 | [0.677-0.690] |
| MrOS-WSC | 0.805 | 0.087 | 0.822 | [0.801-0.809] | 0.699 | 0.126 | 0.722 | [0.693-0.705] |
| SHHS-SSC | 0.816 | 0.090 | 0.839 | [0.812-0.821] | 0.722 | 0.129 | 0.755 | [0.716-0.729] |
| SHHS-WSC | 0.824 | 0.089 | 0.846 | [0.820-0.828] | 0.733 | 0.128 | 0.762 | [0.727-0.739] |
| SSC-WSC | 0.742 | 0.110 | 0.755 | [0.737-0.748] | 0.620 | 0.145 | 0.634 | [0.613-0.627] |
| **3** | | | | | | | | |
| **Overall** | 0.808 | 0.092 | 0.830 | [0.807-0.810] | 0.711 | 0.131 | 0.739 | [0.709-0.713] |
| ISRUC-MrOS-SHHS | 0.820 | 0.092 | 0.844 | [0.815-0.825] | 0.732 | 0.134 | 0.766 | [0.725-0.738] |
| ISRUC-MrOS-SSC | 0.798 | 0.088 | 0.816 | [0.794-0.802] | 0.694 | 0.129 | 0.720 | [0.688-0.700] |
| ISRUC-MrOS-WSC | 0.811 | 0.083 | 0.828 | [0.807-0.815] | 0.711 | 0.119 | 0.735 | [0.705-0.717] |
| ISRUC-SHHS-SSC | 0.807 | 0.090 | 0.828 | [0.803-0.812] | 0.714 | 0.126 | 0.739 | [0.708-0.721] |
| ISRUC-SHHS-WSC | 0.817 | 0.091 | 0.842 | [0.813-0.822] | 0.728 | 0.128 | 0.759 | [0.722-0.735] |
| ISRUC-SSC-WSC | 0.755 | 0.109 | 0.775 | [0.750-0.760] | 0.639 | 0.150 | 0.670 | [0.631-0.646] |
| MrOS-SHHS-SSC | 0.833 | 0.071 | 0.848 | [0.829-0.837] | 0.744 | 0.109 | 0.766 | [0.739-0.750] |
| MrOS-SHHS-WSC | 0.840 | 0.073 | 0.854 | [0.836-0.843] | 0.753 | 0.109 | 0.774 | [0.748-0.759] |
| MrOS-SSC-WSC | 0.795 | 0.088 | 0.811 | [0.791-0.800] | 0.687 | 0.123 | 0.706 | [0.681-0.693] |
| SHHS-SSC-WSC | 0.807 | 0.101 | 0.833 | [0.802-0.812] | 0.710 | 0.142 | 0.744 | [0.703-0.717] |
| **4** | | | | | | | | |
| **Overall** | 0.821 | 0.085 | 0.840 | [0.819-0.823] | 0.728 | 0.124 | 0.755 | [0.726-0.731] |
| ISRUC-MrOS-SHHS-SSC | 0.827 | 0.078 | 0.843 | [0.823-0.831] | 0.739 | 0.115 | 0.764 | [0.733-0.744] |
| ISRUC-MrOS-SHHS-WSC | 0.835 | 0.075 | 0.850 | [0.831-0.838] | 0.747 | 0.112 | 0.768 | [0.742-0.753] |
| ISRUC-MrOS-SSC-WSC | 0.794 | 0.097 | 0.817 | [0.789-0.799] | 0.687 | 0.139 | 0.716 | [0.680-0.694] |
| ISRUC-SHHS-SSC-WSC | 0.819 | 0.091 | 0.843 | [0.814-0.823] | 0.728 | 0.131 | 0.759 | [0.721-0.734] |
| MrOS-SHHS-SSC-WSC | 0.830 | 0.076 | 0.846 | [0.826-0.834] | 0.741 | 0.112 | 0.763 | [0.736-0.747] |

The total number of training records were fixed at $N = 500$ for all configurations. ISRUC: Institute of Systems and Robotics, University of Coimbra Sleep Cohort; MrOS: The Osteoporotic Fractures in Men Sleep Study; SHHS: Sleep Heart Health Study; SSC: Stanford Sleep Cohort; WSC: Wisconsin Sleep Cohort.